%% file: main.tex
\documentclass{article}

\usepackage{microtype}
\usepackage{graphicx}
\usepackage{subfigure}
\usepackage{booktabs} 
\usepackage{amsfonts}       
\usepackage{nicefrac}       
\usepackage{amsmath} 
\usepackage{enumitem,kantlipsum}
\usepackage{float}
\usepackage[export]{adjustbox}
\usepackage{caption}
\usepackage{multirow}
\usepackage{array}
\newcolumntype{P}[1]{>{\centering\arraybackslash}p{#1}}
\usepackage{textcomp}
\usepackage{pifont}
\newcommand{\cmark}{\ding{51}}%
\newcommand{\xmark}{\ding{55}}%
\input{math_commands}

\usepackage{subfigure}
\usepackage{booktabs} 
\usepackage{balance}
\usepackage{hyperref}
\usepackage{url}
\usepackage[utf8]{inputenc} 
\usepackage[T1]{fontenc}    
\usepackage{hyperref}       
\usepackage{url}            
\usepackage{booktabs}       
\usepackage{amsfonts}       
\usepackage{nicefrac}       
\usepackage{wrapfig}

\usepackage{makecell}
\usepackage{todonotes}
\usepackage{mdwlist}

\newcommand{\bx}{\textit{X}}

\newcommand{\bz}{\textit{Z}}

\newcommand{\bT}{\boldsymbol{\theta}}

\newcommand{\pT}{p_{\bT}}

\newcommand{\fT}{\mathcal{F}}

\usepackage{hyperref}

\usepackage[accepted]{icml2021}

\icmltitlerunning{CDCGen: Cross-Domain Conditional Generation via Normalizing Flows and Adversarial Training}

\begin{document}

\twocolumn[
\icmltitle{CDCGen: Cross-Domain Conditional Generation via Normalizing Flows and Adversarial Training}




\begin{icmlauthorlist}
\icmlauthor{Hari Prasanna Das}{to}
\icmlauthor{Ryan Tran}{to}
\icmlauthor{Japjot Singh}{to}
\icmlauthor{Yu-Wen Lin}{to}
\icmlauthor{Costas J. Spanos}{to}
\end{icmlauthorlist}

\icmlaffiliation{to}{Department of Electrical Engineering and Computer Sciences, University of California, Berkeley}

\icmlcorrespondingauthor{Hari Prasanna Das}{hpdas@berkeley.edu}

\icmlkeywords{Machine Learning, ICML}

\vskip 0.3in
]



\printAffiliationsAndNotice{} 

\input{abstract}
\input{introduction}

\input{preliminary}
\input{related_work}
\input{methods}
\input{experiments}
\input{conclusion}
\balance

\bibliographystyle{icml2021}
\bibliography{sample-base}
\input{appendix}
\end{document}

%% file: math_commands.tex
\usepackage{amsmath,amsfonts,bm}

\def\eqref#1{equation~\ref{#1}}

\def\1{\bm{1}}

\DeclareMathAlphabet{\mathsfit}{\encodingdefault}{\sfdefault}{m}{sl}
\SetMathAlphabet{\mathsfit}{bold}{\encodingdefault}{\sfdefault}{bx}{n}

\def\gT{{\mathcal{T}}}



%% file: abstract.tex
\begin{abstract}
    How to generate \textit{conditional} synthetic data for a domain without utilizing information about its labels/attributes? Our work presents a solution to the above question. We propose a transfer learning-based framework utilizing normalizing flows, coupled with both maximum-likelihood and adversarial training. We model a source domain (labels available) and a target domain (labels unavailable) with individual normalizing flows, and perform domain alignment to a common latent space using adversarial discriminators. Due to the invertible property of flow models, the mapping has exact cycle consistency. We also learn the joint distribution of the data samples and attributes in the source domain by employing an encoder to map attributes to the latent space via adversarial training. During the synthesis phase, given any combination of attributes, our method can generate synthetic samples conditioned on them in the target domain. Empirical studies confirm the effectiveness of our method on benchmarked datasets. We envision our method to be particularly useful for synthetic data generation in label-scarce systems by generating non-trivial augmentations via attribute transformations. These synthetic samples will introduce more entropy into the label-scarce domain than their geometric and photometric transformation counterparts, helpful for robust downstream tasks.
\end{abstract}

%% file: introduction.tex
\section{Introduction}
A large majority of the real-world signals obtained are unlabeled, and require significant human effort or machine intelligence for labeling. This has led to a surge in popularity of unsupervised learning algorithms. A prominent branch of such algorithms, generative modeling, has proven to be efficient in transferring knowledge gained from one (or multiple) domain(s) to other domain(s). Variants of such approaches include cross-domain translation~\citep{zhu2017unpaired,isola2017image}, domain adaptation for classification~\citep{hoffman2017cycada,zou2019consensus} etc. By jointly modeling the data samples and their labels/attributes, variations of conditional synthesis methods have been proposed ~\citep{mirza2014conditional,odena2016semi,liu2019conditional}, which during inference phase, can generate synthetic conditional samples. We combine both the above avenues of cross-domain translation along with conditional synthesis and propose a framework capable of generating conditional samples for a domain without utilizing its labels/attributes. 

Prior works on cross-domain translation involve construction of a mapping between two (or more) unpaired domains. The translation consistency is maintained by introducing some form of inductive bias terms such as cycle consistency~\citep{zhu2017unpaired}, semantic consistency~\citep{royer2020xgan}, entropic regulation~\citep{courty2017joint} etc. Most of the proposed models for domain translation are generative adversarial network (GAN)~\citep{goodfellow2014generative} based and involve many-to-one/one-to-many mappings, making the cycle consistency only approximate. A recent work, Alignflow~\citep{grover2019alignflow} achieves exact cycle consistency by modeling the domains with normalizing flows via a common latent space. Normalizing flows~\citep{dinh2016real,kingma2018glow} are a class of generative models which map an unknown and complex data distribution to a latent space with a simple (e.g. standard gaussian) prior distribution via invertible mappings. Another benefit with having flow model mappings is that they offer a rich latent space, which is suitable for a number of downstream tasks, such as semi-supervised learning \citep{odena2016semi}, synthetic data augmentation and adversarial training \citep{cisse2017parseval}, text analysis and model based control etc. 
\begin{table*}[t]
  \caption{Comparison of CDCGen with state-of-the-art cross domain translation and conditional syntheis models. Across the board, CDCGen features all the advantages over other models.}
  \label{tab:comparison}
  \resizebox{\textwidth}{!}{%
  \begin{tabular}{p{4.5cm} | p{2.1cm} | p{2.6cm} | p{3.5cm} | p{4cm}}

    \toprule
    \centering
    Model     & Cross-Domain Translation & Cycle Consistency & Independent Conditional Synthesis & Availability of Latent Space Embeddings \\
    \midrule
    \multicolumn{1}{c|}{XGAN \citep{royer2020xgan}} & \multicolumn{1}{c|}{\cmark} & \multicolumn{1}{c|}{Approximate}  &  \multicolumn{1}{c|}{\xmark} & \multicolumn{1}{c}{\xmark}    \\
    \multicolumn{1}{c|}{CycleGan \citep{zhu2017unpaired}} & \multicolumn{1}{c|}{\cmark} & \multicolumn{1}{c|}{Approximate}  &  \multicolumn{1}{c|}{\xmark} & \multicolumn{1}{c}{\xmark}    \\
    \multicolumn{1}{c|}{\citet{taigman2016unsupervised}} & \multicolumn{1}{c|}{\cmark} & \multicolumn{1}{c|}{Approximate}  &  \multicolumn{1}{c|}{\xmark} & \multicolumn{1}{c}{\xmark}\\
    \multicolumn{1}{c|}{Alignflow \citep{grover2019alignflow}} & \multicolumn{1}{c|}{\cmark} & \multicolumn{1}{c|}{Exact}  &  \multicolumn{1}{c|}{\xmark} & \multicolumn{1}{c}{\cmark}\\
    \multicolumn{1}{c|}{CGAN \citep{mirza2014conditional}} & \multicolumn{1}{c|}{\xmark} & \multicolumn{1}{c|}{--}  &  \multicolumn{1}{c|}{\cmark} & \multicolumn{1}{c}{\xmark}\\
    \multicolumn{1}{c|}{ACGAN \citep{odena2017conditional}} & \multicolumn{1}{c|}{\xmark} & \multicolumn{1}{c|}{--}  &  \multicolumn{1}{c|}{\cmark} & \multicolumn{1}{c}{\xmark}\\
    \multicolumn{1}{c|}{CAGlow \citep{liu2019conditional}} & \multicolumn{1}{c|}{\xmark} & \multicolumn{1}{c|}{--}  &  \multicolumn{1}{c|}{\cmark} & \multicolumn{1}{c}{\cmark}\\
    
    \multicolumn{1}{c|}{\textbf{CDCGen (ours)}} & \multicolumn{1}{c|}{\cmark} & \multicolumn{1}{c|}{\textbf{Exact}}  &  \multicolumn{1}{c|}{\cmark} & \multicolumn{1}{c}{\cmark}\\
    \bottomrule
  \end{tabular}}
\end{table*}
Conditional synthesis has been explored by CGAN~\citep{mirza2014conditional} by augmenting the conditions with the data and processing it via GAN and by ACGAN~\citep{odena2016semi} by introducing an auxilliary classifier for the conditions. This becomes challenging for flow models which are bijective in nature, and hence indirect methods must be adopted to jointly model data and the conditions. ~\citet{liu2019conditional} propose an encoder-discriminator-classifier-decoder based approach on flow latent space which can generate synthetic samples for a domain by passing its conditions via encoders to the data via a flow network. They show improvements in varying the quality of generated images for handles relating to various features.

We present CDCGen, a generative framework that is capable of transferring knowledge across multiple domains and using it to generate synthetic samples for domains lacking information about labels/attributes. We model the label/attribute scarce domain as the target, and a related domain with available information about its labels/attributes as the source. We model the source and target domain via normalizing flows with a common latent space. For conditional synthesis, we introduce a variant of ACGAN by using it on the learned latent space rather than the data space, and train it with only the data and available labels from the source domain. The features can be manipulated easily in the latent space, which is learnt by the conditional synthesis network. During inference phase, CDCGen offers independently specifying conditions, encoding them to a common latent space and moving through the inverse flow to generate conditional synthetic samples in the target domain. Table~\ref{tab:comparison} summarizes the comparison between CDCGen and other related models for different feature availability. CDCGen comes out to be an amalgamation of all features available among the model selections. We establish the CDCGen framework and conduct empirical evaluations with benchmarked image datasets. CDCGen shows encouraging performance in domain alignment, as well as conditional generation for all source and target combinations.

%% file: preliminary.tex
\section{Preliminaries}
Flow-based generative models and generative adversarial networks constitute the major building blocks for proposed CDCGen. The functioning of both the classes of generative models are included in the Appendix.

%% file: related_work.tex
\section{Related Work}
We discuss the related work from two perspectives relevant to the CDCGen framework, namely cross-domain translation and conditional synthesis.
\subsection{Cross-Domain Translation}
Cross-domain translation involves construction of mappings between two or more domains, by training on unpaired data samples in both the domains. Such a problem is under-constrained and involves aligning the domains in feature space via mappings. A number of research in this space~\citep{zhu2017unpaired,royer2020xgan,liu2016coupled,yi2017dualgan,tzeng2017adversarial} introduce a form of cycle consistency loss which ensures that by translating an image from one domain to another domain via mappings and then applying reverse mappings to translate back yields the same image.  XGAN~\citep{royer2020xgan} uses additional loss terms to incorporate semantic consistency across domains, to match the subspace for embedding from multiple domains and prior knowledge via pre-trained models. However, since all the above models involve GAN based architectures, they lack a latent space embedding useful for downstream manipulation tasks ~\citep{kingma2018glow}. Moreover, since the mappings are not guaranteed to be invertible, the cycle consistency is only approximate.

Alignflow~\citep{grover2019alignflow} involves modeling each of the domains via normalizing flow mappings to a common latent space~\citep{dinh2016real,das2019dimensionality,das2019likelihood}. It has a hybrid training objective constituting both maximum likelihood estimation and adversarial training. Moreover, since flows are invertible mappings, Alignflow achieves exact cycle consistency. However, flow models, by virtue of the training procedure, face a challenge to align domains which are apart in terms of semantics and/or style, apparent from the generated samples quality in comparison with GANs. For CDCGen, we use the best of both worlds: flow model mappings for the domains to a common latent space, along with loss terms useful to align the domains in the embedding space. CDCGen offers a rich latent space, which is utilized for conditional synthesis in attribute scarce domains.
\subsection{Conditional Synthesis}\label{sec:related-work-cond-syn}
Conditional generative models have been introduced to generate desired synthetic data by incorporating conditions information in model design. From CGAN ~\citep{mirza2014conditional} which is a modification of conventional GANs and works by feeding the label/attribute information to the generating block, conditional synthesis has seen different algorithmic variations~\citep{hong2018inferring,wang2018high,odena2016semi}. A notable work, ACGAN~\citep{odena2016semi} employs an auxilliary classifier for the discriminator to classify the class labels. A recent work, CAGlow~\citep{liu2019conditional} proposes a variant of ACGAN with an encoder-decoder network, adding ability to model unsupervised conditions. Additionally, above works deal with conditional generation in a single domain. We use a variant of ACGAN over a shared latent space for multiple domains, thereby transferring knowledge from label-rich domains to perform conditional synthesis in label-scarce domains. 

%% file: methods.tex
\section{The CDCGen Framework}
\begin{figure*}[t]
    { \begin{center}                             
        \subfigure[\label{fig:training}CDCGen Training Schematic]{\includegraphics[width=0.52\textwidth]{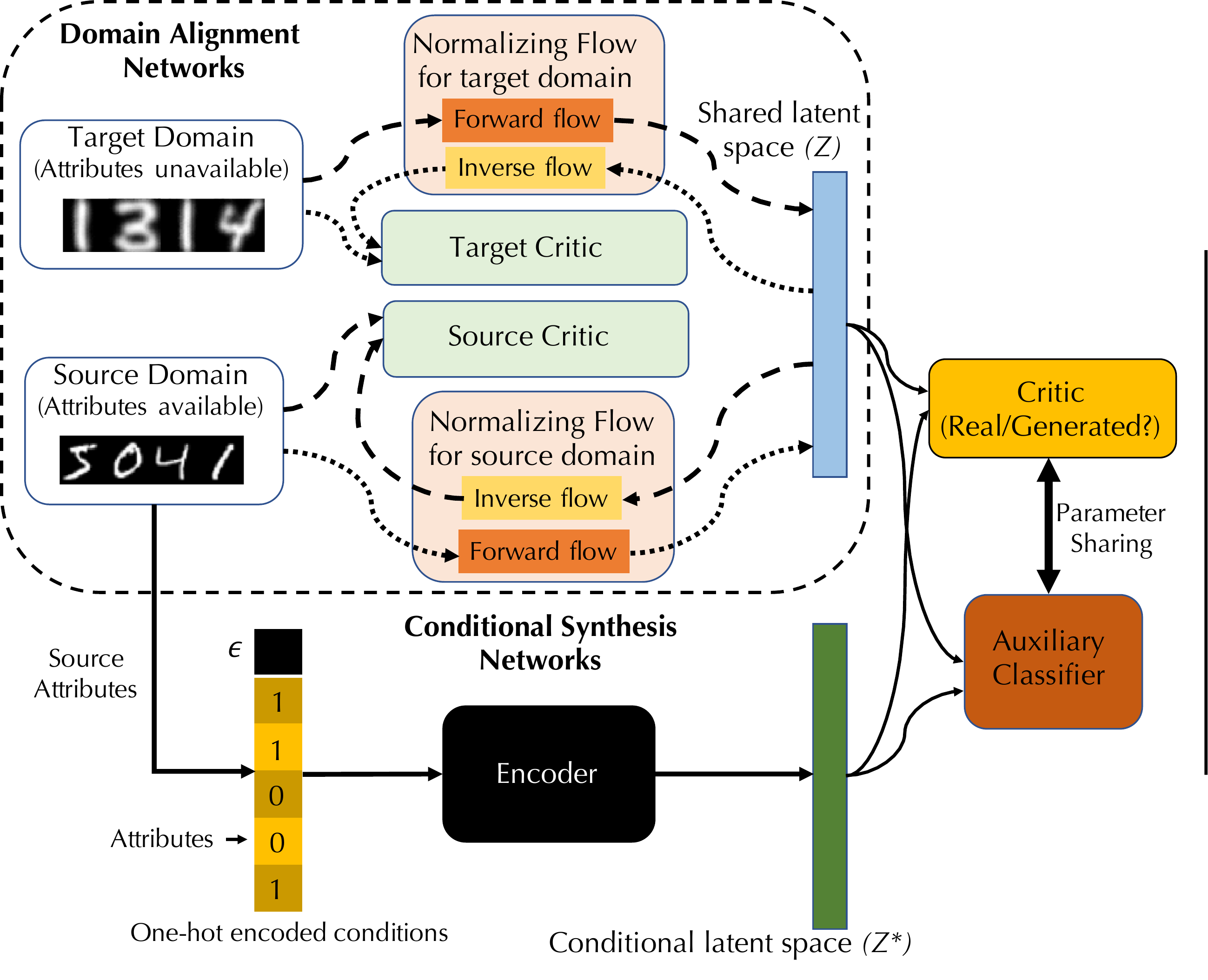}}
        \hspace{0.01\textwidth}
        \subfigure[\label{fig:inference}CDCGen Inference Schematic]{\includegraphics[width=0.39\textwidth]{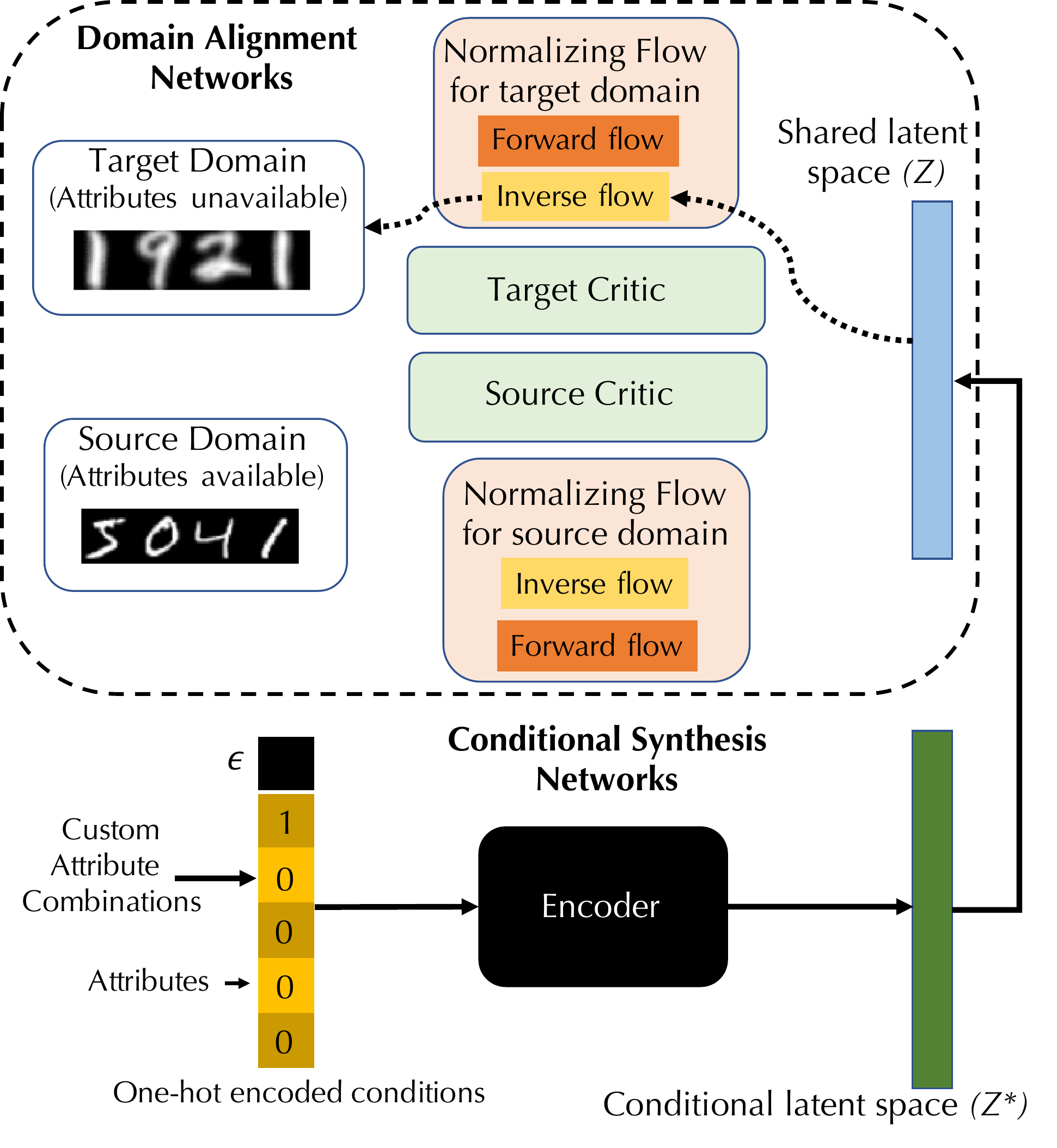}}
     \end{center}
    }
    \caption{Illustration of training and inference methods in CDCGen. The networks inside the dashed box are for domain alignment (Sec~\ref{sec:dom-algn}) and those outside are for conditional synthesis (Sec ~\ref{sec:cond-syn}).}
    \label{fig:illustration}
\end{figure*}

In this section, we will present the CDCGen framework capable of generating conditional synthetic samples for a domain in an unsupervised setting. We select a domain with availability of information about the labels/attributes (namely, the source domain) and has shared attributes with the domain for which we don't have information about labels/attributes (namely, the target domain). Under this setting, the framework consists of two major networks: one for domain alignment and another for conditional synthesis. We consider the case of two domains, but under the assumption of having shared attributes between the source and target domains, the proposed method generalizes to multiple domains seamlessly. 

\subsection{Domain Alignment}\label{sec:dom-algn}
The first step in CDCGen is to align the source and target domains. Let the source and target domain be denoted by $\mathcal{D}_s$ and $\mathcal{D}_t$ with unknown marginal density $p_s$ and $p_t$ respectively. Both the domains are mapped via invertible transformations (normalizing flows) $\mathcal{F}_s$ and $\mathcal{F}_t$ to a common latent space $Z$, which serves as a shared feature space for alignment. We assume the shared latent space follows a normal gaussian distribution $p(z)$, common for training of most of the state-of-the-art flow models. The relationship between the sample space and latent space can be represented as,
\begin{align*}
    \mathcal{D}_s \xrightarrow[]{\mathcal{F}_s} Z \xleftarrow[]{\mathcal{F}_t} \mathcal{D}_t
\end{align*}
Note that the invertible nature of the flow model is helpful in two different ways,

\begin{itemize}
    \item It provides a mechanism to translate between source and target domains, with invertible mappings $\mathcal{F}_{s\rightarrow t} = \mathcal{F}_t^{-1} \circ \mathcal{F}_s$ and $\mathcal{F}_{t\rightarrow s} = \mathcal{F}_s^{-1} \circ \mathcal{F}_t $.
    \item It helps achieve exact cycle consistency (as introduced in CycleGAN ~\citep{zhu2017unpaired} to ensure accurate representation of the mappings)  between the domains, since $\mathcal{F}_{s\rightarrow t} \circ \mathcal{F}_{t\rightarrow s} = \mathcal{F}_t^{-1} \circ \mathcal{F}_s \circ \mathcal{F}_s^{-1} \circ \mathcal{F}_t = I$, where $I$ is the identity matrix.
\end{itemize}
    
We use a hybrid training objective involving both maximum likelihood estimation and adversarial training. Flow models are trained with an unsupervised maximum likelihood loss, with a normal gaussian prior on the latent space $Z$. Since there are two flow models involved for the two domains, the maximum likelihood loss is expressed as,
\begin{align*}
    \mathcal{L}_{MLE}(\mathcal{F}_s) + \mathcal{L}_{MLE}(\mathcal{F}_t)
\end{align*}
For cross-domain mappings, adversarial loss terms are introduced. These terms introduce inductive bias required for cross domain translation \citep{zhu2017unpaired}. We employ critics $\mathcal{C}_s$ and $\mathcal{C}_t$ for source and target domains respectively, which distinguish between real samples (sampled from the same domain) vs. generated samples (obtained via cross-domain mappings). For example, the adversarial loss for source domain can be expressed as,
\begin{align*}
    \mathcal{L}_{ADV}(\mathcal{C}_s,\mathcal{F}_{t\rightarrow s}) = &\mathbb{E}_{x_s \sim p_s}[\log \mathcal{C}_s(x_s)] \\+ &\mathbb{E}_{x_t \sim p_t}[\log (1-\mathcal{C}_s(\mathcal{F}_{t\rightarrow s}(x_t)))]
\end{align*}
We also use a domain-adversarial loss \citep{ganin2016domain} which forces the embeddings learnt by the flow models $\mathcal{F}_s$ and $\mathcal{F}_t$ to lie in the same subspace. This is achieved by training a classifier $\mathcal{C}_{DAL}$ which takes the latent space embeddings for each domain and classifies the sample to be coming from $\mathcal{D}_s$ or $\mathcal{D}_t$. It is trained in an adversarial manner, with a classification loss function $\ell(\cdot,\cdot)$, such as cross-entropy. $\mathcal{L}_{DAL}$ can be expressed as,
\begin{align*}
    \mathcal{L}_{DAL}(\mathcal{F}_s,\mathcal{C}_{DAL}) = \mathbb{E}_{x_s \sim p_s} \ell(\mathcal{D}_s,\mathcal{C}_{DAL}(\mathcal{F}_s(x_s))) 
\end{align*}
Finally, for domain alignment, the overall loss term is,
\begin{align*}
    \mathcal{L}_{Domain\;Alignment}(\mathcal{F}_s,\mathcal{F}_t, \mathcal{C}_s, \mathcal{C}_t, \mathcal{C}_{DAL} ; \lambda_s, \lambda_t, \gamma_s, \gamma_t) =\\ \mathcal{L}_{ADV}(\mathcal{C}_s,\mathcal{F}_{t\rightarrow s}) + \mathcal{L}_{ADV}(\mathcal{C}_t,\mathcal{F}_{s\rightarrow t})  +\gamma_s\mathcal{L}_{DAL}(\mathcal{F}_s,\mathcal{C}_{DAL}) \\+ \gamma_t\mathcal{L}_{DAL}(\mathcal{F}_t,\mathcal{C}_{DAL})-\lambda_s\mathcal{L}_{MLE}(\mathcal{F}_s) - \lambda_t\mathcal{L}_{MLE}(\mathcal{F}_t)
\end{align*}
where, hyperparameters $\lambda_s$ and $\lambda_t$ dictate the relative contribution of maximum likelihood loss, and $\gamma_s$ and $\gamma_t$ correspond to contribution of domain adversarial loss, both as compared to the adversarial loss. The objective is minimized w.r.t.
the parameters of the flow models $\mathcal{F}_s$ and $\mathcal{F}_t$ and maximized w.r.t. parameters of $\mathcal{C}_s$, $\mathcal{C}_t$ and $\mathcal{C}_{DAL}$. This procedure is illustrated in the dashed box in Fig.~\ref{fig:training}.

\begin{figure*}[t]
    { \centering                             
        \subfigure[\label{fig:mnist_to_usps_alignflow}Result with MNIST as source and USPS as target]{\includegraphics[width=0.49\textwidth]{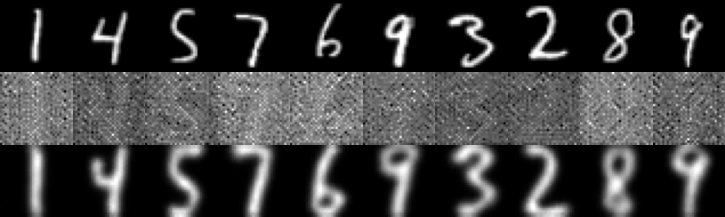}}
        \hspace{0.01\textwidth}
        \subfigure[\label{fig:usps_to_mnist_alignflow}Result with USPS as source and MNIST as target]{\includegraphics[width=0.49\textwidth]{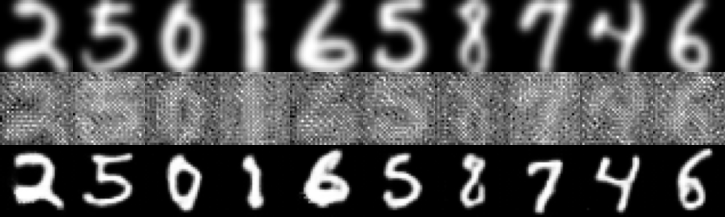}}
    }
    \caption{Results for domain alignment between source and target domains. The top row has original samples from the source domain. The middle row is the corresponding latent space mapping and the bottom row is the sample obtained by translating it to the target domain. The USPS images are slightly blurred due to the upscaling applied as standard pre-processing.}
    \label{fig:mnist_usps_alignflow}
    \centering{
    \includegraphics[width=0.9\textwidth]{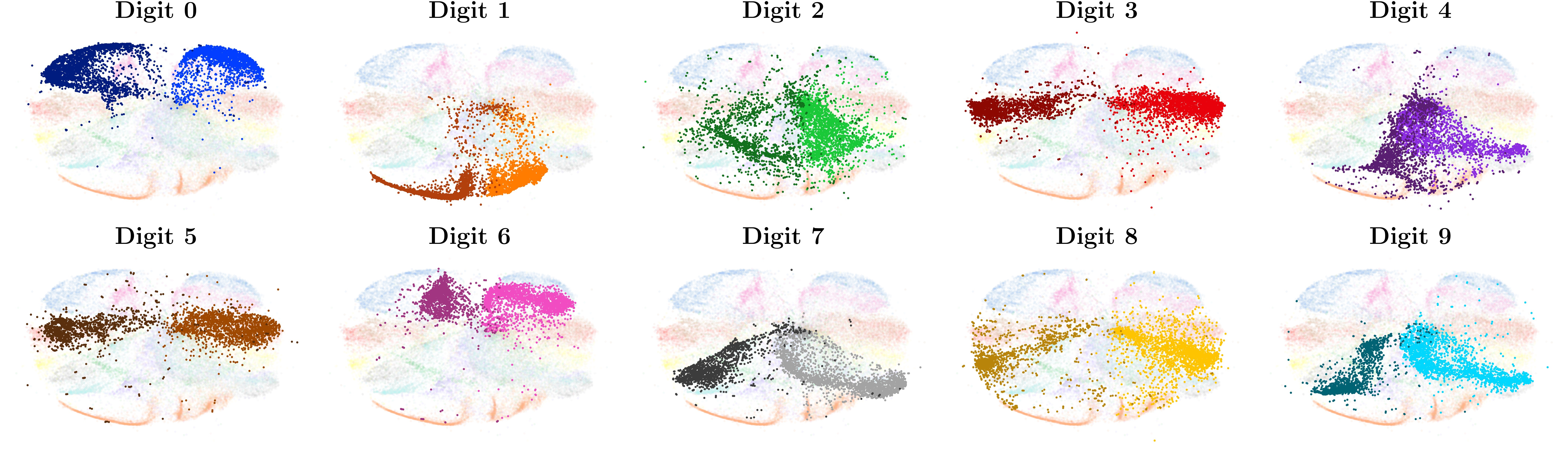}
    \caption{t-SNE representation of shared latent space for MNIST $\leftrightarrow$ USPS. For each digit, points for USPS are visualized with the darker colors, and points with lighter colors correspond to MNIST.}
    \label{fig:tsne_alignflow}}
\end{figure*}

\subsection{Conditional Synthesis}\label{sec:cond-syn}
For conditional synthesis, we propose a variant of ACGAN ~\citep{odena2017conditional}. Instead of using class/attribute conditioning on the sample space as done in ACGAN, we use it in the shared latent space. Under the setting of our problem, we don't have any information about the labels/attributes in the target domain. So, for the conditional synthesis part, only the attributes available from the source domain are used for training. 

We denote the available source attributes/conditions as  $c_s\sim p(c_s)$, represented as one-hot encodings. Our network consists of an encoder to model the conditions, a critic to differentiate between the real and generated latent vectors, and an auxiliary classifier to classify the encoded conditions. We will introduce each of the above components and their associated loss functions separately.

\textbf{Encoder:} An encoder network $E$ encodes the conditions $(c_s,\epsilon)$ into a latent space $Z^*$ (separate from the shared latent space $Z$ for aligned domains), where $\epsilon$ is sampled from standard gaussian distribution $(p(\epsilon))$ and is helpful for incorporating stochastic behavior among condition vectors. Let the distribution for the above mentioned latent space be denoted as $p^*(z)$. Our objective is to minimize the Jensen-Shannon (JS) divergence between the encoded distribution $p^*(z)$ and the shared latent distribution $p(z)$ for aligned domains $\mathcal{D}_s$ and $\mathcal{D}_t$. So, the encoder loss is represented as,
\begin{align*}
    \mathcal{L}_E = \mathbb{E}_{\epsilon \sim p(\epsilon),c_s\sim p(c_s)}[\log\mathcal{C}(E(c_s,\epsilon))]
\end{align*}
where, $\mathcal{C}$ is a critic, more about which we describe now.

\textbf{Critic:} A critic $\mathcal{C}$ discriminates between the latent vectors coming from generated conditional distribution $p^*(z)$ and real shared latent distribution $p(z)$ for aligned domains. This is an adversarial loss which is trained so as it is unable to distinguish the latent vectors at equilibrium, thus enabling the encoder $E$ to generate latent vectors close to the real shared latent distribution $p(z)$. The loss function for $\mathcal{C}$ is,
\begin{align*}
    \mathcal{L}_{CRITIC} = \mathbb{E}_{z\sim p^*(z)}[\log \mathcal{C}(z)] + \mathbb{E}_{z\sim p(z)}[1- \log\mathcal{C}(z)]
\end{align*}

\textbf{Classifier:} A classifier takes the latent vectors ($z\sim p^*(z)$ and $z\sim p(z)$) as input and classifies the conditions ($c_s$). The classifier loss is a cross entropy loss between the predicted and true conditions. If the class posterior probabilities are $q(c_s|z)$, the classifier loss function can be expressed as,
\begin{align*}
    \mathcal{L}_{CLASSIFIER} = &\mathbb{E}_{z\sim p^*(z),c_s\sim p(c_s)}q(c_s|z) \\+ &\mathbb{E}_{z\sim p^(z),c_s\sim p(c_s)}q(c_s|z)
\end{align*}
The overall loss function for the conditional synthesis part is,
\begin{align*}
    \mathcal{L}_{Conditional\;Synthesis} = &\beta_E\mathcal{L}_E + \beta_{Cr}\mathcal{L}_{CRITIC} \\+ &\beta_{Cl}\mathcal{L}_{CLASSIFIER}
\end{align*}
where $\beta_E,\beta_{Cr},\beta_{Cl}$ are hyperparameters. The critic and the classifier networks share their parameters except for their output blocks. Conditional synthesis procedure is illustrated in Fig.~\ref{fig:training}.

\subsection{Inference}
CDCGen can generate conditional samples in the target domain, even when the training process does not utilize its class/attribute information. To generate samples with conditions $\tilde{c}_s$, a latent vector $\tilde{z}$ is generated by encoding the one-hot conditions $\tilde{c}_s$ and $\tilde{\epsilon}\sim p(\epsilon)$ via the encoder network, i.e. $\tilde{z} = E(\tilde{c}_s,\tilde{\epsilon})$. Then the latent vector $\tilde{z}$ is passed via the inverse flow $\mathcal{F}_t^{-1}$ to generate the desired sample in the target domain, i.e. $\mathcal{F}_t^{-1}(\tilde{z})$. The inference schematic is illustrated in Fig.~\ref{fig:inference}.

%% file: experiments.tex
\section{Experiments}

\begin{figure*}[t]
\centering
\subfigure[\label{fig:mnist_to_usps_cdcgen}Generated samples for USPS as target]{\includegraphics[width=0.35\textwidth]{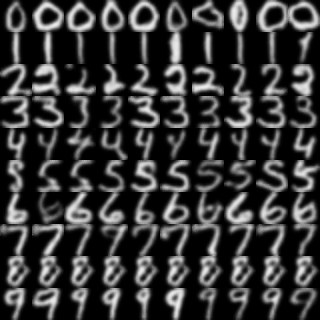}}
         \hspace{0.05\textwidth}
        \subfigure[\label{fig:usps_to_mnist_cdcgen}Generated samples for MNIST as target]{\includegraphics[width=0.35\textwidth]{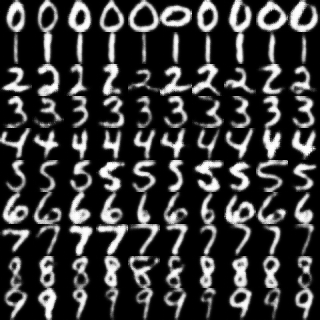}}
    \caption{Conditional synthetic samples generated by CDCGen. The rows represent conditioned digit classes (0-9) and the columns include more samples for each class.}
    \label{fig:mnist_usps_cdcgen}
\end{figure*}

In this section, we empirically evaluate CDCGen for synthetic generation in label scarce domains.

\textbf{Datasets:} We perform experiments on 2 standard image datasets for digits, namely MNIST \citep{lecun1998gradient} and USPS. 
MNIST contains $60,000$ training and $10,000$ test images with ten classes corresponding to digits from $0$ to $9$. USPS has $7291$ training and $2007$ test data with the same classes as MNIST. To address this data imbalance, for each domain, we sample $542$ images from the original training set for each class to form the new training set. To form the test set, we sample $147$ images from the original test sets for each class.  
We resize all the images to $32 \times 32$ for training and synthesis.

\textbf{Source and Target Domain Combinations:} We consider two cases, first with MNIST as the source and USPS as the target domain, and second, with the roles interchanged, i.e. USPS as the source and MNIST as the target. 
We report results for domain alignment and as well as subsequent conditional synthesis in the target domain, all while not using any labels from that domain. 

\textbf{Networks:} We use architecture from RealNVP \citep{dinh2016real} for each of the domain flows ($\mathcal{F}_{s}$ and $\mathcal{F}_{t}$). Typical configurations for RealNVP can be specified as a tuple comprising $N_{scales}$ (number of scales), $N_{channels}$ (number of channels) in the intermediate layers, and $N_{blocks}$ (number of residual blocks in the scaling and translation networks of the coupling layers). For MNIST $\leftrightarrow$ USPS case, both $\mathcal{F}_s$ and $\mathcal{F}_t$ are set to RealNVP($2$, $64$, $8$). The critics ($\mathcal{C}_s$ and $\mathcal{C}_t$) used convolutional discriminators from PatchGAN \citep{isola2017image} , each with $16$ filters in the critic's first convolutional layer. 
For conditional synthesis, we concatenate the one-hot vector of labels with components of random noise as input to the encoder. The vector then passes through one fully-connected layer and eight transposed convolutional layers with upsampling scale $2$, $2$, $2$, $2$, $2$, $1$, $1$, $1$ and channel sizes $256$, $1024$, $512$, $256$, $128$, $64$, $32$, $16$ respectively. The supervision block contains four convolutional layers with stride $2$ and channel sizes $64$, $128$, $256$, $512$. This is followed by two separate fully-connected layers for each network head, one for outputting probabilities of real or fake and the other for classifying the one-hot encoded conditions.


\textbf{Optimizer:} For training the domain alignment network, we use the Adam optimizer with $\beta_1=0.5$, $\beta_2=0.999$, and learning rate $1\cdot10^{-6}$. For training the conditional synthesis network, we use the Adam optimizer with $\beta_1=0.5$, $\beta_2=0.999$, and learning rate $2\cdot10^{-5}$.


\subsection{Domain Alignment}
In this section we present the results for domain alignment between source and target combinations. Fig.~\ref{fig:mnist_to_usps_alignflow} shows the source MNIST samples and corresponding USPS samples by translating it via the forward source and inverse target flows. The middle sample is visualization of corresponding latent space sample. Fig.~\ref{fig:usps_to_mnist_alignflow} depicts the same with USPS as the source and MNIST as the target. It can be observed that the class identity is preserved with the translation with the style adapted for the target domains. The sharpness of the translated samples are compromised, which is a result of the flow model assigning some probability mass to all the samples it is fed. This is unlike pure GAN models which selectively assign probability mass to meaningful samples.

Another interesting observation is the appearance of digit class identity in latent space visualizations. This is particularly useful from the perspective of CDCGen, since the conditional synthesis network works based on the latent space mappings from both the domains. 

We present the t-SNE embeddings for the shared latent space in our proposed domain alignment network for MNIST and USPS in Fig.~\ref{fig:tsne_alignflow}. It can be observed that the visualization has distinct clusters for each digit class, but the embeddings from both the source and target domain are close and belong to the same cluster for the overall digit class clustering. The visualization allows us to infer that the latent space has learned a subspace corresponding to each digit, and interpolating across this subspace is effectively a conditional feature-preserving domain transfer.
\vspace{-2mm}
\subsection{Conditional Synthesis}
We trained the conditional synsthesis part of CDCGen (Section~\ref{sec:cond-syn}) with source labels to generate conditional synthetic samples in the target domain. Fig.~\ref{fig:mnist_to_usps_cdcgen} shows the samples generated with USPS as the target domain and Fig.~\ref{fig:usps_to_mnist_cdcgen} shows the samples generated with MNIST as the target. Each row corresponds to the digit classes which are assigned as conditions. It can be observed that CDCGen is able to generate synthetic samples belonging to the digit class as conditioned. There are also variations among the samples across different columns which shows the stochastic nature of generation by CDCGen. The compromise in sharpness of the samples generated is owed from the domain alignment mappings by flow models.
\vspace{-2mm}

%% file: conclusion.tex
\section{Conclusions}
In this work, we proposed CDCGen, a generative framework capable of generating conditional synthetic samples for domains without the requirement of obtaining its labels/attributes. 
We also conducted empirical studies with standard image datasets to observe feature transfer and independent conditional generation. We are working on CDCGen implementations for datasets with complex interactions between features, e.g. facial data. In the future, making the conditional generation models across multiple domains can be studied with varying levels of label availability (few-shot learning) for target domain. CDCGen can also be adapted for other modalities including audio and tabular data. It can also be used alongside real-world applications ~\citep{zou2019wifi,zou2019machine,konstantakopoulos2019design,chen2021enforcing,periyakoil2021environmental,das2019novel,das2020occupants,liu2018personal,liu2019personal,donti2021machine,jin2018biscuit} where having access to diverse conditional data is important, but is hard to obtain, hence the need for synthetic data.

%% file: appendix.tex
\appendix
\section{Functioning of Flow and GAN based generative models}
\subsection{Flow-based Generative Models}
Let $\bx$ be a high-dimensional random vector with unknown true distribution $p(x)$. The following formulation is directly applicable to continuous data, and with some pre-processing steps such as dequantization \citep{uria2013rnade,salimans2017pixelcnn++,ho2019flow++} to discrete data. Let $\bz$ be the latent variable with a known standard distribution $p(z)$, such as a standard multivariate gaussian. Using an i.i.d. dataset $\mathcal{D}$, the target is to model $\pT(x)$ with parameters $\bT$. A flow, $\fT$ is defined to be an invertible transformation that maps observed data $\bx$ to the latent variable $\bz$. A flow is invertible, so the inverse function $\gT$ maps $\bz$ to $\bx$, i.e.
\begin{align}
    \bz = \fT(\bx) = \gT^{-1}(\bx)\;\;\text{and}\;\;
    \bx = \gT(\bz) = \fT^{-1}(\bz)
\end{align}
The log-likelihood can be expressed as,
\begin{align}
\log\pT(x) &= \log p(z) + \log\left|\det\left(\frac{\partial \fT(x)}{\partial x^{T}}\right)\right|\label{eq:flow_func_repr}
\end{align}
where $\cfrac{\partial \fT(x)}{\partial x^{T}}$ is the Jacobian of $\fT$ at $x$. The training of flow models is accomplished via maximum-likelihood estimation. Confirming with the qualifying properties for a flow as above, different types of flow models have been introduced to efficiently estimate the distribution density and generate synthetic samples \citep{kingma2018glow,dinh2016real,dinh2014nice,chen2019residual}.
\subsection{Generative Adversarial Networks (GANs)}
GANs \citep{goodfellow2014generative} are a class of implicit generative models which work based on the principles of a mini-max game. It involves a generator $\mathcal{G}$ which is tasked to generate synthetic samples from standard noise distribution and a critic $\mathcal{C}$ which learns to discriminate the samples generated by $\mathcal{G}$ and samples from original data distribution $p_{data}$. The training objective for a GAN is given by,
\begin{align*}
    \min_{\mathcal{G}}\max_{\mathcal{C}} \mathcal{L}(\mathcal{C},\mathcal{G}) = &\mathbb{E}_{x \sim p_{data}}[\log \mathcal{C}(x)] \\+ &\mathbb{E}_{z \sim \text{standard noise distribution}}[\log (1-\mathcal{C}(\mathcal{G}(z)))]
\end{align*}
At nash equillibrium, the generator and critic are optimal, and the generator is capable of generating samples resembling original data. GAN based models have been particularly successful in generating high-fidelity images \citep{karras2018style,brock2018large}, manipulating features of images to generate custom samples ~\citep{radford2015unsupervised}, audio generation ~\citep{engel2019gansynth,donahue2018adversarial}, video generation~\citep{tulyakov2018mocogan} etc. Despite the potential of GANs in generating qualitative samples, they are hard to train due to the mini-max optimization. Unlike flow models, they lack a latent space suitable for a number of downstream applications. Another major problem with GANs is mode collapse, where the generator starts producing the same output (or a small set of outputs) over and over again. A number of remedies have been proposed to tackle this over the years \citep{arjovsky2017wasserstein,salimans2016improved,metz2016unrolled}.